\documentclass[conference]{IEEEtran}
\IEEEoverridecommandlockouts
\usepackage{cite}
\usepackage{amsmath,amssymb,amsfonts}
\usepackage{algorithmic}
\usepackage{graphicx}
\usepackage{textcomp}
\usepackage{xcolor}

\usepackage{multirow}
\usepackage{multicol}
\usepackage{enumitem}
\usepackage{bm} 
\usepackage{booktabs}
\usepackage{subfigure}
\usepackage{stfloats}
\setlist{leftmargin=3.5mm}
\usepackage[raggedrightboxes]{ragged2e}

\def\BibTeX{{\rm B\kern-.05em{\sc i\kern-.025em b}\kern-.08em
    T\kern-.1667em\lower.7ex\hbox{E}\kern-.125emX}}
\begin{document}

\title{Enhancing Graph Neural Networks with Limited Labeled Data by Actively Distilling Knowledge from Large Language Models}


\author{\IEEEauthorblockN{Quan Li$^{1}$, Tianxiang Zhao$^{1}$, Lingwei Chen$^{2}$, Junjie Xu$^{1}$, Suhang Wang$^{1}$}
\IEEEauthorblockA{
$^1$\textit{Pennsylvania State University},
University Park, PA, USA \\
$^2$\textit{Wright State University}, Dayton, OH, USA\\
\{qbl5082, tkz5084, junjiexu, szw494\}@psu.edu, lingwei.chen@wright.edu}

}

\maketitle

\begin{abstract}
Graphs are pervasive in the real-world, such as social network analysis, bioinformatics, and knowledge graphs. Graph neural networks (GNNs) have great ability in node classification, a fundamental task on graphs. Unfortunately, conventional GNNs still face challenges in scenarios with few labeled nodes, despite the prevalence of few-shot node classification tasks in real-world applications. To address this challenge, various approaches have been proposed, including graph meta-learning, transfer learning, and methods based on Large Language Models (LLMs). However, traditional meta-learning and transfer learning methods often require prior knowledge from base classes or fail to exploit the potential advantages of unlabeled nodes. Meanwhile, LLM-based methods may overlook the zero-shot capabilities of LLMs and rely heavily on the quality of generated contexts. In this paper, we propose a novel approach that integrates LLMs and GNNs, leveraging the zero-shot inference and reasoning capabilities of LLMs and employing a Graph-LLM-based active learning paradigm to enhance GNNs' performance. Extensive experiments demonstrate the effectiveness of our model in improving node classification accuracy with considerably limited labeled data, surpassing state-of-the-art baselines by significant margins.

\end{abstract}


\section{Introduction}

Graphs have become increasingly recognized as one of the powerful data structures to perform real-world content analysis \cite{chen2019semi,
mohamed2020social,reau2023deeprank}. They are adept at representing complex relationships and uncovering hidden information between objects across various domains. Among various tasks on graphs, node classification stands out as a classic task with broad applications, such as sentiment analysis \cite{li2021dual} and user attribute inference \cite{li2022distilling}. Recently, graph neural networks~\cite{kipf2016semi,abu2019mixhop,hamilton2017inductive} have shown great power in node classification. Generally, GNNs adopt the message-passing mechanism, which updates a node's representation by aggregating its neighbors' information, facilitating the implicit propagation of information from labeled nodes to unlabeled ones. This strategy has substantially enhanced performance across various benchmark datasets \cite{wu2020comprehensive}.


Despite the great success of GNNs in node presentation learning and node classification, they often struggle to generalize effectively when labeled data is scarce. However, in many real-world applications, due to various reasons such as labeling cost and privacy issues, one often needs to train a GNN classifier with sparse labels, which is known as few-shot node classification. For example, labeling a large number of web documents can be both costly and time-consuming \cite{wei2021few,han2018fewrel}; similarly, in social networks, privacy concerns limit access to personal information, leading to a scarcity of attribute labels~\cite{li2022distilling}. 
Consequently, when confronted with such datasets, GNNs may exhibit poor generalization to unlabeled nodes.
To tackle the few-shot learning problem, various methods have been proposed, such as meta-learning~\cite{finn2017model,ding2021few,ding2022meta}, transfer learning~\cite{wan2021contrastive,zhu2021transfer}, and adversarial reprogramming \cite{chen2022adversarially}. However, they still require a substantial amount of labeled nodes in each class to achieve satisfactory results \cite{yao2021functionally} or require auxiliary labeled data to provide supervision. 

Recently, Large Language Models (LLMs) have demonstrated their outstanding generalizability in zero-shot learning and reasoning \cite{chen2023label,kojima2022large}. 
Several efforts have been taken in introducing LLMs to graph learning, such as pre-processing of textual node attributes or taking textual descriptions of rationales as inputs \cite{liu2023one,he2023explanations}, leveraging LLMs to construct graph structure \cite{guo2024graphedit,jiang2023structgpt}, and generating new nodes \cite{yu2023empower}. For example, Chen et al. \cite{chen2023label} first leveraged LLMs as annotators to provide more supervision for graph learning. Yu et al. \cite{yu2023empower} leveraged the generative capability of LLMs to address the few-shot node classification problem. These works demonstrate that LLMs can enhance GNNs from different perspectives. However, they typically treat LLMS merely as annotators or generators for node classification tasks, overlooking their untapped potentials, such as the capacity to uncover hidden insights within the results and their zero-shot reasoning ability, which could significantly enhance GNNs' performance for few-shot learning tasks. 

In this paper, we introduce a novel few-shot node classification model that enhances GNNs' capabilities by actively distilling knowledge from LLMs. Unlike previous approaches, our model uses LLMs as ``teachers'', capitalizing on their zero-shot inference and reasoning capabilities to bolster the performance of GNNs in few-shot learning scenarios. However, there are two primary challenges: (i) LLMs cannot consistently deliver accurate predictions for all nodes. How to select nodes that LLMs can provide high-quality labels that can benefit GNN most; and (ii) How to effectively distill the knowledge from LLMs to GNNs. To address these challenges, we propose an active learning-based knowledge distillation strategy that selects valuable nodes for LLMs and bridges the gap between LLMs and GNNs. This approach significantly enhances the efficacy of GNNs when labeled data is scarce.
We first explore the metrics that affect the correctness of LLMs' predictions. Then, we employ LLMs as a teacher model and leverage them to perform on the limited training data, generating soft labels for training nodes along with logits and rationales. These outputs are used to supervise GNNs in learning from two perspectives: probability distribution and feature enhancement at the embedding level. In this way, GNNs can learn the hidden information from unlabeled nodes and the detailed explanation provided by LLMs. Furthermore, we introduce a novel Graph-LLM-based active learning approach to establish a connection between LLMs and GNNs, which effectively select nodes for which GNNs fail to provide accurate pseudo-labels but LLMs can offer reliable pseudo-labels, thereby enabling GNNs to leverage the zero-shot capabilities of LLMs and enhance their performance with limited data. Afterward, the selected pseudo-labels are merged with the true labels to train the final few-shot node classification model. Our major contributions are: 
\begin{itemize}

    \item We innovate a semi-supervised learning model by distilling knowledge from Large Language Models and leveraging the enhanced rationales provided by Large Language Models to help GNNs improve their performance. 

    \item We design and implement a Graph-LLM-based active learning paradigm to enhance the performance of GNNs. This is achieved by identifying nodes for which GNNs struggle to generate reliable pseudo labels, yet LLMs can provide dependable predictions, which leverage the zero-shot 
    ability of LLMs to enhance the performance of GNNs. 

    \item Extensive experiments on various benchmark datasets demonstrate the effectiveness of our proposed framework for node classification tasks with limited labeled data. 
\end{itemize}

\section{Related Work}


\noindent\textbf{Graph Neural Networks} Graph Neural Networks (GNNs) have garnered widespread attention for their effective exploitation of graph structure information. There are two primary types of GNNs: spectral-based and spatial-based. Kipf and Welling~\cite{kipf2016semi} followed the idea of CNNs and proposed the Graph Convolutional Network (GCN) to aggregate information within the spectral domain using graph convolution. Different from GCN, Graph Attention Network (GAT)~\cite{velivckovic2017graph} and GraphSAGE~\cite{hamilton2017inductive} emerged as spatial-based approaches. GAT applies the attention mechanism to learn the importance of the neighbors when aggregating information. GraphSAGE randomly samples the number of neighbors of a node and aggregates information from these local neighborhoods. Despite their extensive application across various domains, GNNs often face challenges due to limited labeled data. Existing convolutional filters or aggregation mechanisms struggle to effectively propagate labels throughout the entire graph when only few labeled data points are available~\cite{li2018deeper}.


\noindent\textbf{Few-shot Node Classification} In real-world graph learning tasks, obtaining high-quality labeled samples can be particularly challenging due to various factors such as the high cost involved in annotation processes or the limited access to node information. Thus, researchers proposed different methods to improve the performance of GNNs with only few labeled data. Most recent advancements in few-shot node classification (FSNC) models have mainly developed from two approaches: metric-learning and meta-learning. Metric-learning models aim to learn a task-invariant metric applicable across all tasks to facilitate FSNC \cite{hao2019collect,jiang2020multi}. Prototypical network \cite{snell2017prototypical} and relation network \cite{sung2018learning} are two classic examples, where the former uses the mean vector of the support set as a prototype and calculates the distance metrics to classify query instances, and the latter trains a neural network to learn the distance metric between the query and support set. Meta-learning models use task distributions to conduct meta-training, learning shared initialization parameters that are then adapted to new tasks during meta-testing \cite{ding2021few,zhou2019Meta,g-meta}. These approaches have demonstrated effectiveness compared to metric-learning, which often struggles due to task divergence issues. However, meta-learning requires significant amounts of data for meta-training, sourced from the same domain as meta-testing, thereby severely limiting its practical applicability.
\textit{Different from metric-learning and meta-learning models, in this paper, we propose to distill knowledge from LLMs to GNNs, leverage the LLMs' zero-shot ability and reasoning ability to improve GNNs for few-shot node classification}.


\noindent\textbf{LLMs for Text-Attributed Graphs} Recently, LLMs have garnered widespread attention and experienced rapid development, emerging as a hot topic in the artificial intelligence area. 
Within the graph domain, LLMs show their generalizability in dealing with Text-Attributed Graphs (TAGs). Chen et al. \cite{chen2023label} demonstrated the power of LLMs' zero-shot ability on node classification tasks. Moreover, LLMs also demonstrate their power in providing rationales to enhance node features \cite{he2023explanations} and construct edges in graphs \cite{sun2023large}. Liu et al. \cite{liu2023one} further proposed OFA to encode all graph data into text and leverage LLMs to make predictions on different tasks. Despite their remarkable proficiency in understanding text, LLMs still face limitations when it comes to processing graph-structured data. Therefore, leveraging LLMs' zero-shot ability and integrating them with GNNs has emerged as the latest state-of-the-art approach in text-attributed graph learning \cite{chen2023label}.


\noindent\textbf{Active Learning} Active learning (AL) \cite{shen2017deep, bachman2017learning, cai2017active, gao2018active, wu2019active} is a widely adopted approach across various domains for addressing the issue of label sparsity. The core concept involves selecting the most informative instances from the pool of unlabeled data. Recently, many works \cite{cai2017active, gao2018active, hu2020graph} integrate GNNs with AL to improve the representative power of graph embeddings. However, how to leverage AL to build connections between LLMs and GNNs and improve the performance of GNNs has emerged as a problem. Chen et al. \cite{chen2023label} first leverage active learning to select nodes that are close to the cluster center under the label-free setting and use LLM as an annotator to create labels for these nodes. However, their approach simply leverages LLMs to annotate nodes and ignores the benefits of unlabeled nodes and the zero-shot reasoning ability of LLMs. And, under the few-shot setting, GNNs themselves can provide relatively high-quality pseudo-labels for those nodes close to the cluster center, which waste resources if we use LLMs to generate pseudo-labels for those nodes. Moreover, prior research primarily concentrated on selecting data with the highest confidence score during the AL process. In our work, instead of focusing on nodes where GNNs have high confidence, we prioritize nodes where GNNs struggle to provide pseudo-labels with high confidence scores but LLMs can provide reliable predictions. This approach is motivated by our integration of LLMs as teacher models to enhance the performance of GNNs by leveraging LLMs' zero-shot pseudo-labeling and reasoning ability. Through active learning, we integrate LLMs into GNNs, enabling LLMs to instruct GNNs with data that GNNs find challenging to label confidently.

\section{Preliminaries}
\label{sec:preliminary}
In this section, we conduct preliminary experiments to reveal the metrics that can affect LLMs to generate high-quality pseudo labels and formulate the problem.


\noindent\textbf{Notations} We use $\mathcal{G}=(\mathcal{V}, \mathcal{E})$ to denote a graph, where $\mathcal{V} = \{v_1, v_2, \dots, v_N\}$ is a set of $N$ nodes and $\mathcal{E}$ is a set of edges. We use $\mathbf{A}$ to denote the adjacency matrix, where $\mathbf{A}_{ij} = 1$ means nodes $v_i$ and $v_j$ are connected; otherwise $\mathbf{A}_{ij}=0$. The text-attributed graph can be defined as $\mathcal{G}_T = (\mathcal{V}, \mathbf{A}, \mathcal{X})$, where $\mathcal X = \{\mathcal{X}_1, \mathcal{X}_2, \cdots, \mathcal{X}_N\}$ denotes the set of raw texts and can be encoded as text embeddings $\mathbf{X} = \{\mathbf{x}_1, \mathbf{x}_2, \cdots, \mathbf{x}_N\}$. In semi-supervised learning, The node set $\mathcal{V}$ can be divided into two different sets: (1) the labeled node set $\mathcal{V}_l$ and (2) the unlabeled node set $\mathcal{V}_u$. Moreover, we use $\mathcal{V}_S$ to denote the labeled node set including both original labeled data and the data selected through active learning, and use $\mathcal{Y}$ to represent the label set, where $\mathcal{Y} = \{y_1,y_2,\cdots,y_N\}$. 

\begin{figure}[t]
    \centering


    \includegraphics[clip, trim= 20 0 15 5, width = \linewidth]{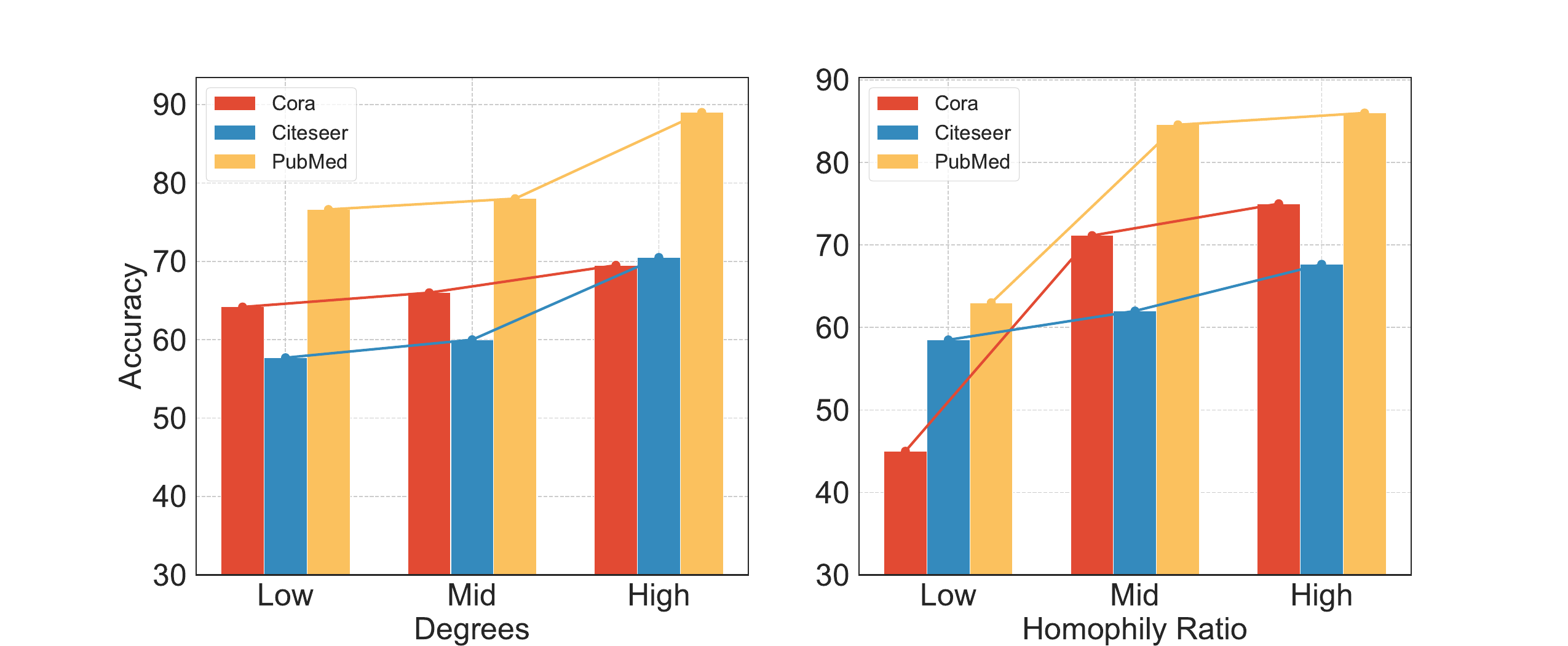}
    \vspace{-2em}
    \caption{Preliminary experiments: the impact of the degrees (left) and homophily ratio (right) to LLMs}
    \label{fig:prexp}
    \vskip -1.5em
\end{figure}

\subsection{Understanding LLM's Capability} 
As the sparse label challenges GNN, in this paper, we aim to imbue GNNs with the zero-shot learning prowess of LLMs, thereby elevating their performance in scenarios with limited labeled data. 
However, LLMs might be good at classifying certain nodes while performing poorly on other nodes. Thus, it is important to identify nodes that LLMs can provide superior pseudo-labels with rationales, whereas GNNs cannot, which can better enhance GNNs' performance. Hence, we first conduct preliminary experiments to understand key factors pivotal for LLMs in generating reliable pseudo-labels.

LLMs may benefit from various metrics to perform node classification well. Particularly, in graph $\mathcal{G}$, certain metrics exert a more pronounced influence on the correctness of LLM predictions on nodes, which include: 
1) the degree of a node, and 2) the homophily ratio. 
Both degrees and the homophily ratio are important for a node. The former indicates how many nodes will be affected by a node, while the latter suggests that the node tends to connect with others having similar features. Therefore, it's crucial to examine how the degree and the homophily ratio affect the performance of LLMs' predictions. 

We conduct preliminary experiments to understand how these factors influence the classification performance of LLMs. Specifically, we use the following equation to compute the homophily ratio: $\mathbf{HR} = \frac{\text{\#  of neighbors have same label}}{\text{total  \#  of neighbors}}$.

We divide degree and homophily into 3 categories: highest, middle, and lowest,
and select 200 nodes for each category. Specifically, We sort the nodes based on the degrees and the homophily ratio in descending order, evenly selecting 200 nodes from the head, tail, and middle of the node list for the highest, lowest, and middle categories, respectively. The GPT-3.5-turbo is used for testing. We provide the raw text \(\mathcal{X}_i\) and the potential classes to the LLMs, asking them to assign a label from the given class to \(\mathcal{X}_i\). Then, we compare the results from LLM and the ground truth labels for evaluation. 

Figure \ref{fig:prexp} shows the preliminary experimental results. From the figure, 
we can find that the performance of LLMs can be affected by the degree and the homophily ratio of a node. When we use the nodes with more indegrees or higher homophily ratio, the classification accuracy of LLMs is increased significantly. Compared with degrees, LLMs are more sensitive to the homophily ratio. From the right figure of Figure \ref{fig:prexp}, it is obvious that the performance of LLMs changed drastically across nodes with different homophily ratios. For example, the accuracy of LLMs for nodes with the lowest homophily ratio in the Cora dataset is around $40\%$, but it reaches $75\%$ on nodes with the highest homophily ratio. This is because nodes with more degrees and a higher homophily ratio tend to be closer to the distribution center and occupy well-connected positions in the graph. These nodes exert indirect influence and are naturally associated with richer textual information, making them more significant and representative within clusters. For instance, in citation networks, papers with more citations are more representative and easily distinguishable within their field. These representative and distinct contexts lead LLMs to achieve better classification outcomes.


Our preliminary experiments demonstrate that LLMs are capable of generating high-quality pseudo-labels for nodes with a higher homophily ratio and more degree, which paves us a way to effectively select nodes to query LLMs to obtain high-quality knowledge for enhancing GNNs knowledge. 


\begin{figure*}[t]
    \centering

    \includegraphics[scale = 0.5]{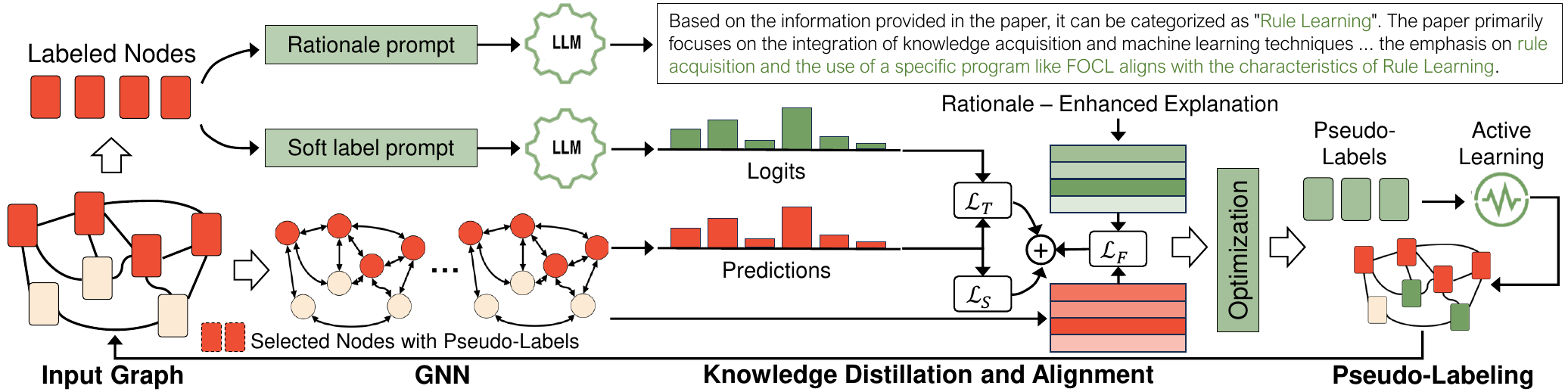}
    \vskip -1em
    \caption{An illustration of the proposed framework}\label{fig:overview}
    \vskip -1em   
\end{figure*}

\subsection{Problem Statement}
As LLMs could not give reliable knowledge for all the nodes, in this paper, we study a novel problem of how to effectively leverage LLMs to enhance the performance of few-shot node classification over graphs. Given a text-attributed graph $\mathcal{G}_T = (\mathcal{V}, \mathbf{A}, \mathcal{X})$ with a very limited labeled node set $\mathcal{V}_l$ (i.e. $|\mathcal{V}_l| \ll |\mathcal{V}_u|$) and their label set $\mathcal{Y}_l$, a budget size $B$ (note that the budget size $B$ is the number of nodes per class), and a large language model $LLM$, we aim to train a GNN that can have better performance with only few available labeled nodes by querying $LLM$ within the budget $B$. 

\vspace{-0.2cm}
\section{Proposed Model}
\label{sec:model}


Though GNNs have shown great power in node classification, the vanilla GNNs suffer from low generalizability with few labeled data for training~\cite{feng2020graph}. 
Thus, in order to enhance the generalizability of GNNs, we propose a framework that integrates GNNs with LLMs and employs a novel Graph-LLM-based active learning strategy to actively distill knowledge from LLMs. Our proposed model uses GNN as the backbone model and takes advantage of LLMs' zero-shot pseudo-labeling and reasoning capabilities, especially for nodes that are difficult for GNN to give accurate predictions. In these instances, LLMs can offer reliable pseudo-labels and provide enhanced rationales, thereby improving the few-shot learning capability of GNNs from distinct perspectives.

An illustration of the framework is shown in Figure~\ref{fig:overview}. 
Specifically, LLM serves as a teacher model, instructing the student model (GNNs) from two distinct perspectives: (1) it imparts the ``correct'' answers to the student model along with the probability distribution for all potential categories, drawing upon its vast knowledge, which teaches GNNs with the output logits; and (2) it explains the rationale behind its decision-making process, providing insights into why certain decisions are made, which serves as the feature teacher to teach GNNs at embedding level. Then the knowledge obtained from LLMs will be distilled to GNNs, and GNNs propagate label information to all unlabeled nodes. We leverage Graph-LLM-based active learning to identify nodes that GNNs struggle to generate reliable pseudo labels but LLMs can provide reliable predictions. These selected nodes are then added to the train set with pseudo labels, and fed to LLMs for logits and rationales, which can further enhance the capability of GNNs under the guidance of LLMs. Finally, we train the ultimate student model, enhancing its generalizability with limited data. Next, we introduce each component in detail.

\subsection{Base GNN Classifier}
\label{subsec:semi-supervise}

As GNNs have shown great power in semi-supervised node classification, we adopt Graph Neural Networks (GNNs) as the backbone models, which can be used to capture the structure information between entities and naturally propagate the information to all unlabeled nodes efficiently. 
We first use SBERT~\cite{reimers2019sentence} to encode raw texts $\mathcal{X}$ to text embeddings $\mathbf{X}$. Then, we use GNNs to perform on the given graph and these embeddings. 
Specifically, the GNN takes the graph $\mathcal{G}_T$ as input and learns the node representation as 
\begin{equation}
    \mathbf{H}^\mathtt{f} = GNN(\mathbf{A}, \mathbf{X})
\end{equation}
where $\mathbf{H}^{\mathtt{f}}$ is the node representation matrix from the last layer of GNN. The final prediction results can be computed as:
\begin{equation} \label{eq:output}
    \mathbf{Z} = \text{softmax} (\mathbf{H}^{\mathtt{f}})
\end{equation}
where $\mathbf{Z}$ is the probabilities for all nodes in the graph. The loss function for training the GNN will be introduced in \ref{subsec:optimize}.

\subsection{Obtaining Knowledge from LLM}
\label{subsec:learningwithLLM}

Despite GNNs showing success in dealing with graph data, the generalizability of GNNs with few available data is still limited. To tackle this challenge, we introduce LLMs as teacher models, leveraging their zero-shot ability \cite{kojima2022large} to instruct GNNs in classification tasks and provide insights into the reasoning behind these decisions. In this way, GNNs can learn hidden label distribution information and enhanced feature information from LLMs, which empower the capabilities of GNNs with scarce labeled data 

To effectively distill knowledge from LLMs to GNN, we consider two types of knowledge: (1) soft labels and logits; and (2) rationales behind LLMs' decision-making process. Soft labels and logits reveal hidden distribution information for unlabeled data, while rationales contribute richer node information. This combination allows GNNs to benefit from the unlabeled data and get enhanced node features. 
We prompt the prediction and reasoning in a two-step manner: first, we input the raw texts into the LLMs to generate the soft labels and logits with the probability distribution. We then let LLMs explain the reason why they make these decisions. Examples of prompts are shown in Table \ref{tab:promot}. To avoid deviations in output formats with multi-task prompts, we use separate prompts for logits and rationales, respectively. Next, we give the details.

\begin{table}[t] 
    \centering
    \caption{Prompt examples} \label{tab:promot}
    \vspace{-1em}
    \footnotesize
    \begin{tabular}{p{0.2\linewidth} | p{0.7\linewidth}}
    \hline
        For soft labels and logits & Paper: $<$Paper Information$>$. Task: 
        For the following categories: $<$categories$>$, which categories does this paper belong to? Provide your $<$k$>$ best guesses within the given categories: $<$categories$>$ and a confidence score that each is correct (0 to 1). The sum of all confidence should be 1. Outputs must be in the given categories. For example:  ``answer'': $<$your first answer$>$, ``confidence'': $<$confidence for first answer$>$, ...  \\
        \hline
         For rationales & Paper: $<$Paper Information$>$. Task: For the following categories: $<$categories$>$, which categories does this paper belong to? Think step by step. Explain your decision in detail. \\ 
         \hline
    \end{tabular}
    \vspace{-1.5em}
\end{table}

\subsubsection{Soft Labels and Logits Generation} For a node $v_i$, we first feeds raw text $\mathcal{X}_i$ into LLMs to generate soft labels $\bar{y}_i$ for $\mathcal{X}_i$ and the logits $\mathbf{l}_i$ for all possible categories. An example of the prompt for soft labels and logit generation is shown in the first row of Table~\ref{tab:promot}. We leverage the zero-shot ability of LLMs to generate relevant reliable soft labels and logits so that GNNs can leverage the hidden information of unlabeled data with knowledge distillation. This can be written as
\begin{equation}
    \bar{y}_i, \mathbf{l}_{i} = LLM(\mathcal{X}_{i};prompt)
    \label{eq:softlabel}
\end{equation}

\subsubsection{Rationales for Feature Enhancement} Traditional knowledge distillation methods primarily utilize the soft labels and logits from the teacher model. Nonetheless, incorporating the rationales behind text decisions can significantly enhance the learning capabilities of GNNs \cite{he2023explanations}. In this context, GNNs are able to learn more informative features from the LLM at the embedding level. 
Consequently, we introduce LLMs as a feature teacher, guiding GNNs to assimilate more informative features in their decision-making process. Unlike previous works that concatenate the enhanced embeddings and the node embeddings or simply replace the node representations directly, we will use a loss function to minimize the difference between them, which will help GNNs learn the enhanced representation while retaining the original node representations. The loss function will be detailed in Section \ref{subsec:knowledgedistillation}.

For a node $v_i$, LLMs will output the classification result for $\mathcal{X}_i$ with a detailed explanation of the decision-making process. The enhanced explanation $\mathcal{R}_{i}$ can be represented as follows:
\begin{equation}
    \mathcal{R}_{i} = LLM(\mathcal{X}_{i};prompt)
    \label{eq:rationalpseudolabels}
\end{equation}
An example of the prompt for rationales is shown in the second row of Table~\ref{tab:promot}. 
Since the rationales we get from LLMs are all textual explanations, we further need to transform them into the embedding level to teach GNNs the more informative features. We use a pre-trained language model such as Sentence BERT (SBERT)~\cite{reimers2019sentence} to get the embeddings for $\mathcal{R}_{i}$, which can be represented as follows:
\begin{equation}
    \bar{\mathbf{r}}_{i} = LM(\mathcal{R}_{i})
    \label{eq:rationalesembedding}
\end{equation}
where $\bar{\mathbf{r}}_{i}$ means the embedding of $i$-th rationale.

However, since the dimensionality of $\bar{\mathbf{r}}_{i}$ may be different from the dimension of the final layer of GNNs, alignment between these representations is necessary. 
While min/max pooling can effectively reduce dimensionality for alignment purposes, it tends to lose information during the pooling process. To retain the enriched information from these rationales, we train a Multi-Layer Perceptron (MLP) using text embeddings $\mathbf{X}_l$ of the limited labeled node set $\mathcal{V}_l$ and their corresponding ground truth labels $\mathcal{Y}_l$, applying the cross-entropy loss function.
This MLP is tasked with aligning the representations between the rationales $\bar{\mathbf{r}}$ and the outputs $\mathbf{H}^{\mathtt{f}}$ of the final layer of GNNs, ensuring that valuable information is retained throughout the alignment process. 
The final representation for $i$-th rationale is generated as follows:
\begin{equation}
    \label{eq:MLPAlign}
    \mathbf{r}_i = MLP(\bar{r}_i)
\end{equation}
where $r$ is the embedding that has the same dimension as the final layer's outputs $\mathbf{H}^{\mathtt{f}}$ in GNNs.

\subsection{Distilling Knowledge to GNN}
\label{subsec:knowledgedistillation}


With the knowledge from LLM represented as $\mathbf{r}_i$ and $\mathbf{l}_i$, we use knowledge distillation~\cite{hinton2015distilling} to distill this knowledge into GNNs. 
Through this process, GNNs can tap into the hidden information behind unlabeled nodes by using output logits to enhance their performance. Moreover, they can achieve improved node representations by incorporating the rationales generated from LLMs to further enrich the depth and quality of the information being processed. Specifically, LLMs serve as a pre-trained teacher model to teach the student model (GNNs) from two distinct perspectives: 1). soft labels and the probability distribution (logits) and 2). the rationales at the embedding level. 

\subsubsection{Loss for Knowledge Distillation} Let $\mathcal{V}_S$ be the set of nodes including the original training data and the data selected through active learning (to be introduced in Section~\ref{subsec:activelearning}). Following~\cite{li2022distilling}, for each $v_i \in \mathcal{V}_s$, we first convert the logits $\mathbf{l}_i$ from LLM as:
\begin{equation} \label{eq:probability}
    p(y_i=j | LLM) = \frac{\exp{\left(\mathbf{l}_{ij}/{\tau}\right)}}{\sum_{c=1}^C \exp{\left(\mathbf{l}_{ic}/{\tau}\right)}}
\end{equation}
where $C$ is the number of classes, $\tau$ is the knowledge distillation (KD) temperature to control how much of the teacher's knowledge is distilled to student model 
and $\mathbf{l}_{ij}$ is the $j$-th elements of $\mathbf{l}_i$. Then, the student can learn the distilled knowledge from the teacher by optimizing the following loss:
\begin{equation} \label{eq:kdloss}
    {\mathcal L}_T = -\frac{1}{|{\mathcal V}_{S}|}\sum_{v_i \in {\mathcal V}_{S}} \sum_{j=1}^C p(y_i=j|LLM)\log Z_{ij}
\end{equation}
where $Z_{ij} $ is the probability that $v_i$ belongs to class $j$ by GNN. This enhances the model's capacity to get insights from unlabeled data and augment its overall learning capabilities.

\subsubsection{Loss for Feature Alignment} We also introduce rationales to augment the node representation from a feature perspective at the embedding level. With $\mathbf{r}$ we get from \ref{subsec:learningwithLLM}, Mean Square Error (MSE) is used to calculate the loss between the rationales and node embeddings $\mathbf{H}^{\mathtt{f}}$ at the final layer of GNNs for all the nodes in the current training set ${\mathcal V}_{S}$ as:
\begin{equation}
    \label{eq:featureloss}
    {\mathcal L}_F =  \frac{1}{|{\mathcal{V}}_{S}|} \sum_{i=1}^{|{\mathcal{V}}_{S}|} ( \mathbf{H}^{\mathtt{f}}_i - \mathbf{r}_i)^2
\end{equation}
where $\mathbf{H}^{\mathtt{f}}_i$ is the node embedding of $v_i$ from GNN. By employing this approach, the GNNs learn informative rationales from LLMs, enhancing their learning capabilities from a feature perspective at the embedding level.

\subsection{Objective Function of Proposed Framework}
\label{subsec:optimize}
The student model itself computes training loss between predictions and labeled or pseudo-labeled data as
\begin{equation} \label{eq:studentloss}
    {\mathcal L}_S = -\frac{1}{|{\mathcal V}_{S}|}  \sum_{v_i \in \mathcal{V}_S} \sum_{c=1}^C \mathbb{I}(y_i=c) \log Z_{ic}
\end{equation}
where $\mathbb{I}$ is the indicator function which outputs 1 if $y_i=c$ otherwise 0. $\mathcal{V}_S$ is the set of labeled or pseudo-labeled nodes, and $y_i$ is the label or pseudo-label of $v_i$.

With the knowledge distillation from LLM, the final loss function of our proposed model can be formalized as follows:
\begin{equation} \label{eq:loss}
    \min_{GNN}{\mathcal L} = (1-\alpha-\beta){\mathcal L}_S + \alpha {\mathcal L}_T + \beta {\mathcal L}_F
\end{equation}
where $\alpha$ and $\beta$ are both balance parameters that are set up to adjust the relative weight of knowledge distillation loss and feature embedding loss, respectively.

\subsection{Graph-LLM-based Active learning}
\label{subsec:activelearning}

To further improve GNNs' few-shot learning ability, we introduce a novel Graph-LLM-based active learning strategy to select valuable nodes for querying LLMs and add them to the training set iteratively. We seek to select $B$ nodes for each class where GNNs exhibit low confidence in classification results, yet LLMs can offer high-quality pseudo-labels based on their inherent knowledge. Through iterative selection, we progressively enhance the GNNs' capabilities. 

As indicated by the preliminary experiment results presented in Section \ref{sec:preliminary}, LLMs demonstrate the ability to generate high-quality pseudo-labels for nodes with higher homophily ratios and more degrees. Thus, we define an evaluation metric that combines the confidence score of GNN's prediction, homophily ratio, and degrees to evaluate if the node in the unlabeled node set $\mathcal V_u$ is valuable for our proposed model. The evaluation metric is defined as follows:
\begin{equation} \label{eq:metricscore}
    \mathbf{S}_{GL_i} = \mathbf{RS}(\mathbf{p}_i) + \mathbf{RS}(HR_i) + \mathbf{RS}(D_i)
\end{equation}
where $\mathbf{S}_{GL_i}$ means the evaluation score for $i$-th node,  $\mathbf{p}_i$ (i.e. $\mathbf{p}_i = \max (\mathbf{Z}_i)$), $HR_i$, and $D_i$ denote the final output confidence score, the homophily ratio, and degree for $i$-th node, respectively. The homophily ratio is calculated using labels generated from GNN. 
The $\mathbf{RS}$ represents a ranking function used to calculate scores for each evaluation metric. Specifically, we arrange the nodes in ascending order according to the evaluation metric results, excluding $\mathbf{p}_i$, and assign scores ranging from 0 to 1 with a step of $1/|\mathcal{V}_u|$. Note that $\mathbf{RS}$ assigns scores to the $\mathbf{p}_i$ in descending order, prioritizing nodes for which GNNs cannot generate reliable pseudo-labels.



Considering the fact that some nodes can better contribute to label propagation and model improvement in the graph, we would like to add a metric to evaluate the importance of the node and to facilitate selecting the most valuable pseudo-labels. Here, we utilize neighborhood entropy reduction to assess the importance of a given node $v_i$ \cite{wang2024distribution}. Specifically, for each node $v_i$ in the candidate set $\mathcal{V}_c$, we compute the entropy reduction in the neighbors' softmax vectors by removing $v_i$ from the node set $\mathcal{V}_n$, which contains the $v_i$ and its neighbors. The basic intuition is that a node is more informative when it can greatly change uncertainty (entropy) within its neighborhood. In other words, the more changes in entropy, the more important a node is. Then we rank and assign scores to these nodes based on the change of entropy. The score of entropy change is defined as follows:
\begin{equation} \label{eq:entropyscore}
    \mathbf{S}_{E_i} = \mathbf{RS}\left(h(\hat{y}_{\mathcal{V}_n - v_i}) - h(\hat{y}_{\mathcal{V}_n}) \right)
\end{equation}
where ${S}_{E_i}$ is the score of entropy change for $v_i$, $h(\cdot)$ is the entropy function, and $\hat{y}$ denotes the pseudo-labels from GNNs ($\hat{y}_{\mathcal{V}_n}$ and $\hat{y}_{\mathcal{V}_n - v_i}$ represent the pseudo labels for the nodes in the set $\mathcal{V}_n$ with and without $v_i$), which is computed based on nodes' logits vectors and the activation function. Thus, the final evaluation metric for $v_i$ is:
\begin{equation} \label{eq:evaluation}
    \mathbf{S}_{i} = \mathbf{S}_{GL_i} + \mathbf{S}_{E_i}
\end{equation}

In each stage, we select subsets of valuable nodes with high $\mathbf{S}_{i}$, each consisting of $b$ nodes per class. We query LLM to obtain the pseudo-label, logits, and rationals. We then add these nodes to the label set and retrain our model using Eq.~\ref{eq:loss}. We continue this process until the total number of nodes meets the budget size $B$ times the number of classes $C$. Here, $B$ is a relatively small budget size, achieving a balance between the cost of querying LLMs and the resultant performance improvement. Finally, the selected nodes with pseudo-labels are used to train the final GNN model. This approach makes GNNs benefit from the various abilities of LLMs, enhancing their performance with scarce labeled data.

\section{Experimental Results and Analysis}

In this section, we present the evaluation results of our proposed few-shot node classification model on benchmark datasets. We aim to answer the following research questions:
\begin{itemize}
    \item \textbf{RQ1:} How does our proposed model perform compared with state-of-the-art baselines under consistent settings?


    \item \textbf{RQ2:} How do different hyper-parameters impact the performance of our model?



    \item \textbf{RQ3:} How do different components in our proposed model contribute to the performance?
\end{itemize}

\subsection{Experimental Setup}
\subsubsection{Datasets} We evaluate our proposed model using three public citation datasets: Cora, Citeseer, and PubMed \cite{kipf2016semi}. These datasets are among the most commonly utilized citation network datasets for evaluating GNN models in node classification tasks. In these datasets, nodes are papers with topics serving as labels. Edges depict citation links between papers, and node features are derived from the title and abstract of papers. The dataset statistics are shown in Table \ref{tab:Dataset_info}.

\begin{table}[t]
    \centering
    \footnotesize
    \caption{Statistics of the datasets}
    \vspace{-0.3cm}
    \tabcolsep=12pt
    \begin{tabular}{cccc}
    \toprule
         \textbf{Dataset} & \textbf{\# nodes} & \textbf{\# edges} & \textbf{\# classes}\\ 
         \hline
         Cora & 2,078 & 5,429 & 7 \\
         Citeseer & 3,327 & 9,228 & 6 \\
         PubMed & 19,717 & 88,651 & 3 \\
         \bottomrule
         
    \end{tabular}
    \label{tab:Dataset_info}
    \vspace{-0.4cm}
\end{table}

\begin{table*}[t]
    \centering
    \footnotesize
    \tabcolsep=18pt
    \caption{Few-shot node classification performance comparison on Citeseer. The best results are highlighted in \textbf{bold}}
    \vskip -1em
    \begin{tabular}{l|cccc}
    \hline
        \multirow{2}{*}{\textbf{Models}} & \multicolumn{4}{c}{\textbf{Citeseer}}\\
        \cline{2-5}
        & 1-shot & 3-shot & 5-shot & 7-shot \\
        \hline
        GCN & 43.72±(1.22) & 52.24±(1.04) & 56.69±(4.39) & 58.20±(3.20)  \\
        GAT & 29.02±(1.35) & 33.91±(1.77) &	36.33±(2.28) &  38.77±(2.74) \\
        GraphSAGE & 43.64±(2.60) & 54.33±(3.20)&	57.76±(3.90) &	58.11±(3.36)\\
        \hline
        MVGRL & 46.13±(3.05) &	55.61±(2.14)&	59.56±(2.56)&	60.52±(3.06)\\
        Meta-PN& 31.00±(4.89)&	42.84±(4.70)&	47.83±(4.30)&	52.33±(3.33)\\
        CGPN & 34.75±(2.79)&	41.69±(1.81)&	45.88±(3.24)&	46.80±(3.32)\\
        \hline
        LLM-based Model & 44.63±(1.72)&	55.75±(0.63)&	58.65±(1.22)&	59.53±(2.39)\\
        \hline
        Our Model (GAT, GPT) & 32.86±(2.29)&	39.51±(1.37)&	41.55±(2.05)&	42.87±(1.84)\\
        Our Model (GraphSAGE, GPT) & 45.41±(3.70)&	56.63±(3.43)&	59.87±(3.73)&	60.87±(3.49)\\
        Our Model (GCN, LLama) & 45.11±(1.82) & 56.22±(2.59) & 60.23±(3.89) & 62.05±(2.64) \\
        Our Model (GCN, Gemini) & 47.04±(2.75) & 57.01±(3.07) & 60.02±(3.19) & 62.55±(3.48)\\
        Our Model (GCN, GPT) & \textbf{47.74±(1.69)}&	\textbf{57.43±(3.33)}&	\textbf{62.19±(2.80)}&	\textbf{63.32±(2.64)}\\

    \hline
    \end{tabular}
    \vspace{-0.1cm}
    \label{tab:compbaselinesCiteseer}
\end{table*}

\begin{table*}[t]
    \centering
    \footnotesize
    \tabcolsep=18pt
    \caption{Few-shot node classification performance comparison on Core. The best results are highlighted in \textbf{bold}}
    \vspace{-0.3cm}
    \begin{tabular}{l|cccc}
    \hline
        \multirow{2}{*}{\textbf{Models}} & \multicolumn{4}{c}{\textbf{Cora}}\\
        \cline{2-5}
        & 1-shot & 3-shot & 5-shot & 7-shot \\
        \hline
        GCN & 48.63±(5.49)&	66.27±(5.04)&	72.15±(2.24)&	74.98±(0.97)\\
        GAT & 41.14±(7.24)&	62.07±(6.96)&	65.27±(3.40)&	68.26±(1.29)\\
        GraphSAGE & 49.81±(4.66)&	64.94±(6.89)&	69.80±(2.41)&	73.37±(1.19)\\
        \hline
        MVGRL & 29.02±(2.49)&	39.60±(3.17)&	41.46±(4.11)&	47.34±(6.01)\\
        Meta-PN& 36.59±(4.66)&	54.55±(3.37)&	59.74±(4.58)&	66.89±(2.66)\\
        CGPN & 47.69±(6.97)&	55.39±(5.01)&	60.57±(2.05)&	62.62±(2.86)\\
        \hline
        LLM-based Model & 48.96±(3.57)&	68.45±(4.80)&	72.65±(1.92)&	74.36±(1.35)\\
        \hline
        Our Model (GAT, GPT) & 47.40±(2.49)&66.66±(3.02) &	68.47±(3.26)&	70.97±(1.77)\\
        Our Model (GraphSAGE, GPT) & 51.55±(2.15)&	68.23±(3.76)&	72.54±(2.40)&	75.39±(0.44)\\
        Our Model (GCN, LLama) &52.57±(1.72)&	73.07±(2.23)&	73.12±(2.84)&	77.26±(2.04) \\
        Our Model (GCN, Gemini) &53.35±(3.68)&	73.42±(1.83)&	75.44±(1.03)&	79.61±(1.02)\\
        Our Model (GCN, GPT) & \textbf{53.82±(2.26)}&	\textbf{74.32±(1.79)}&	\textbf{78.13±(1.63)}&	\textbf{79.73±(1.08)}\\
        
    \hline
    \end{tabular}
    \vspace{-0.1cm}
    \label{tab:compbaselinesCora}
\end{table*}

\begin{table*}[t]
    \centering
    \footnotesize
    \tabcolsep=18pt
    \caption{Few-shot node classification performance comparison on PubMed. The best results are highlighted in \textbf{bold}}
    \vspace{-0.3cm}
    \begin{tabular}{l|cccc}
    \hline
        \multirow{2}{*}{\textbf{Models}} & \multicolumn{4}{c}{\textbf{PubMed}}\\
        \cline{2-5}
        & 1-shot & 3-shot & 5-shot & 7-shot \\
        \hline
        GCN & 54.55±(3.06)&	62.64±(0.85)&	66.53±(3.25)&	69.99±(1.31)  \\
        GAT & 49.53±(2.87)&	59.35±(0.80)&	59.95±(2.22)&	64.58±(1.74) \\
        GraphSAGE & 54.62±(0.02)&	60.58±(2.82)&	63.79±(1.53)&	68.15±(1.67)\\
        \hline
        MVGRL & 41.54±(7.91)&	51.96±(4.74)&	52.14±(4.35)&	54.56±(2.19)\\
        Meta-PN& 40.13±(0.50)&	45.72±(4.38)&	51.47±(3.84)&	56.70±(5.47)\\
        CGPN & 42.73±(3.48)&	39.79±(5.86)&	41.67±(5.74)&	50.02±(3.33)\\
        \hline
        LLM-based Model & 56.45±(0.74)&	61.89±(2.47)&	67.47±(4.30)&	71.25±(2.01)\\
        \hline
        Our Model (GAT, GPT) & 52.77±(3.21)&	62.06±(1.47)&	63.41±(0.88)&	67.78±(2.95)\\
        Our Model (GraphSAGE, GPT) & 57.37±(1.66)&	63.40±(2.71)&	65.98±(3.36)&	71.00±(1.55)\\
        Our Model (GCN, LLama) &57.11±(1.16)&	63.16±(1.25)&	68.00±(3.00)&	73.00±(3.74) \\
        Our Model (GCN, Gemini) &57.29±(2.22)&	64.01±(0.85)&	69.24±(2.93)&	73.05±(2.56)\\
        Our Model (GCN, GPT) & \textbf{58.17±(1.13)}&	\textbf{65.40±(0.84)}&	\textbf{69.37±(2.47)}&	\textbf{73.43±(3.36)}\\
        
    \hline
    \end{tabular}
    \vspace{-0.4cm}
    \label{tab:compbaselinesPubMed}
\end{table*}

\subsubsection{Implementation} Following the traditional dataset split setting, we divide the dataset into three parts: $60\%$ for training, $20\%$ for validation, and $20\%$ for testing. From the training set, we then randomly select $n$-shot samples (i.e. $n\times C$) to be used as training data. It is important to note that since we randomly select $n$-shot nodes as training nodes for the few-shot setting, the choice of seeds will influence the quality of the initial nodes, thereby impacting the classification performance. Hence, we conduct experiments with different seeds $[0, 1, 2]$ and use the average accuracy as our final results.  For our Graph-LLM-based active learning strategies, we set the budget size $B= 3$
, indicating the selection of 3 samples per class in total during the graph active learning process. The balanced parameters are configured as $\alpha = 0.3$ and $\beta = 0.1$, and the KD temperature $\tau = 3$ is used for distilling knowledge from LLMs to GNNs. We use GPT3.5-Turbo as our base LLM model. Additionally, we assess the impacts of different backbone models (GCN, GAT, and GraphSAGE), different LLM base models (GPT3.5-Turbo, Gemini-1.5-flash, and LLama-3.1-8b), different training sizes $\mathbf{N}$, the sample size per class $B$ for active learning, the balance parameter $\alpha$ and $\beta$, and the KD temperature $\tau$ in \ref{subsec: baselines} and \ref{subsec:evaluation}. Note that for the LLM base models, we use the API for GPT and Gemini, while LLama is run locally.

\subsubsection{Baselines}
We use 7 state-of-the-art models as baselines: 3 backbone GNN models (GCN \cite{kipf2016semi}, GAT \cite{velivckovic2017graph}, and GraphSAGE \cite{hamilton2017inductive}), 2 GNN based few-shot learning models (Meta-PN \cite{ding2022meta}, CGPN \cite{wan2021contrastive}), 1 graph self-supervised model (MVGRL \cite{hassani2020contrastive}), and 1 LLM-based few-shot learning model \cite{yu2023empower}.
\begin{itemize}
    \item \textbf{GCN:} GCN conducts convolution operations on graph-structured data, which aggregates information from the neighbors to iteratively update node representations.
    \item \textbf{GAT:} GAT incorporates an attention mechanism into GNN for feature aggregation, which allows GAT to focus on more important neighbors and get better node representations.
    \item \textbf{GraphSAGE:}  GraphSAGE samples neighbors and employs mean aggregation to learn node embeddings, efficiently capturing the graph's structural information.
    \item \textbf{MVGRL:} MVGRL is a benchmark in GNN self-supervised learning by using data augmentation to create diverse views for contrastive learning, employing graph diffusion, and subgraph sampling to enhance its performance.
    \item \textbf{CGPN:} CGPN introduces the concept of poison learning and utilizes contrastive learning to propagate limited labels across the entire graph efficiently.
    \item \textbf{Meta-PN:} Meta-PN uses meta-learning and employs a bi-level optimization to generate high-quality pseudo-labels.
    \item \textbf{LLM-based model:} This LLM-based model leverages LLMs to generate the pseudo nodes for each class, uses LM to encode these nodes and uses an MLP to build edges. 
\end{itemize}
Note that the LLM-based model does not provide the original code; therefore, we independently developed the model based on the paper. For fair comparison, the hyperparameters of all the models are tuned on the validation set. All experimental results are conducted under consistent settings.

\subsection{Comparison with Baselines}
\label{subsec: baselines}
We conduct comprehensive experiments to evaluate the performance of our proposed model compared with 7 state-of-the-art baseline models under consistent settings and aim to answer \textbf{RQ1}. Specifically, we compare our model with seven state-of-the-art models using shots $n = [1, 3, 5, 7]$. Additionally, we assess our model with different backbone models (GCN, GAT, GraphSAGE) and LLMs (GPT, Gemini, LLama). The experiment results can be found in Table \ref{tab:compbaselinesCora}, Table \ref{tab:compbaselinesCiteseer}, and Table \ref{tab:compbaselinesPubMed}. From the experimental results, we can observe that the LLM-based few-shot model either matches or surpasses both traditional GNN models and meta-learning-based models. Obviously, our proposed model with GCN outperforms all state-of-the-art baselines by a large margin in different shots. For example, with the 3-shot setting, the improvement margin of accuracy is $(6 \sim 35)\%$ for Cora, $(2 \sim  23)\%$ for Citeseer, and  $(5 \sim  20)\%$ for PubMed. Compared to the LLM-based few-shot model, our model performs better while requiring fewer detailed outputs, thus reducing costs and resource usage. These observations highlight that our proposed model can achieve state-of-the-art performance with fewer labeled nodes, rendering it a promising approach for few-shot node classification tasks.

From the tables, we can also observe that both the backbone models and LLMs influence the performance of our proposed model. Our model tends to perform better when paired with a high-performing backbone model. Similarly, if the LLM excels in text inference tasks, our model's performance improves accordingly. Despite this dependency, our model significantly outperforms the backbone models alone, highlighting its ability to effectively distill knowledge from LLMs by leveraging their enhanced rationales and soft labels with few labeled data.

\subsection{Hyper-parameter Sensitivity Analysis}
\label{subsec:evaluation}
\begin{figure*}[t]
    \centering

    \subfigure[Alpha]{\includegraphics[width = 0.45\linewidth]{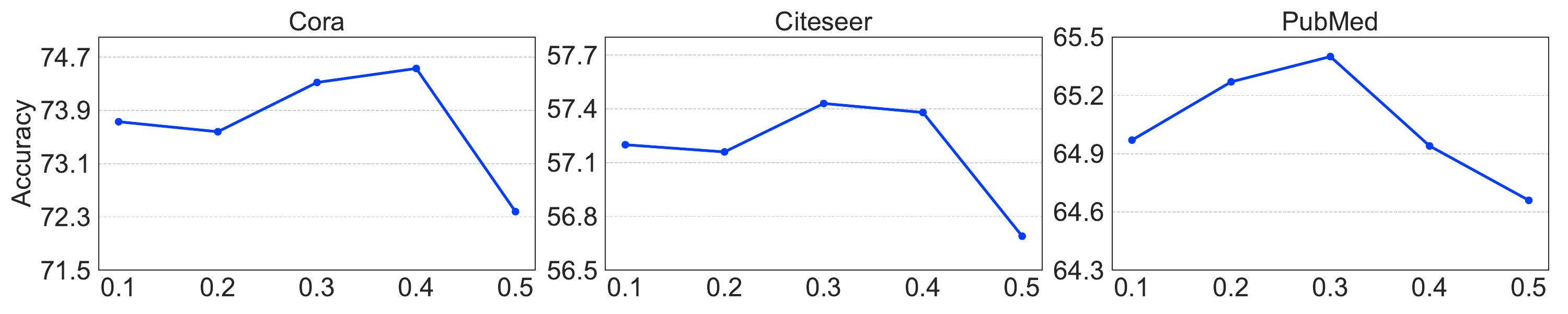} \label{subfig:alpha}} \quad
    \subfigure[Beta]{\includegraphics[width = 0.45\linewidth]{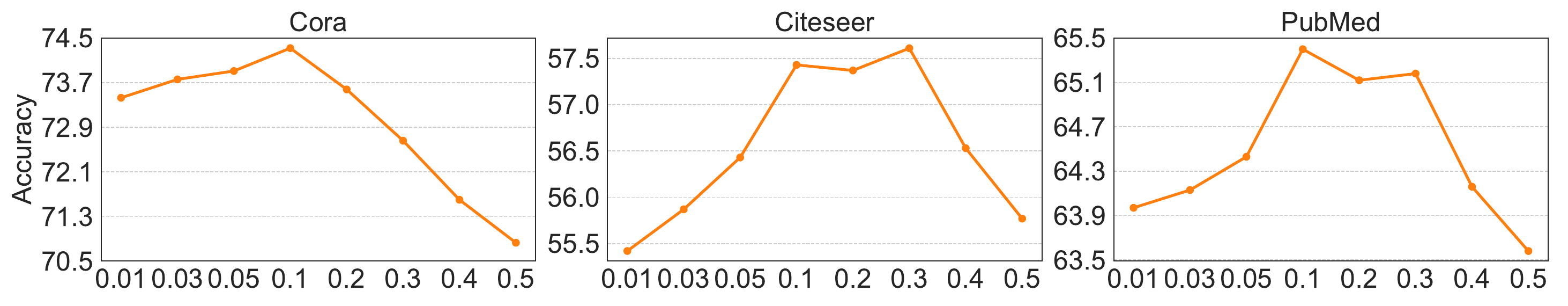} \label{subfig:beta}} \\ 
    \vspace{-0.4em}
    \subfigure[Temperature]{\includegraphics[width = 0.45\linewidth]{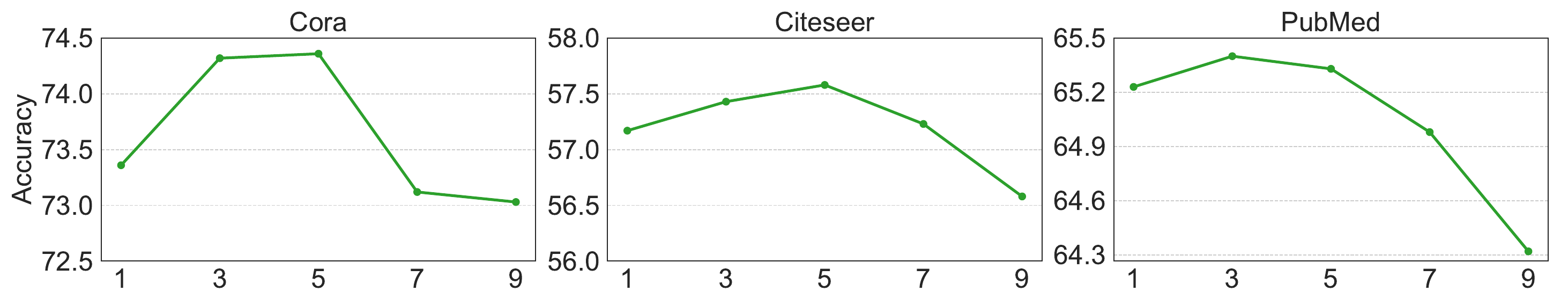} \label{subfig:temperature}}\quad
    \subfigure[Budget Size]{\includegraphics[width = 0.45\linewidth]{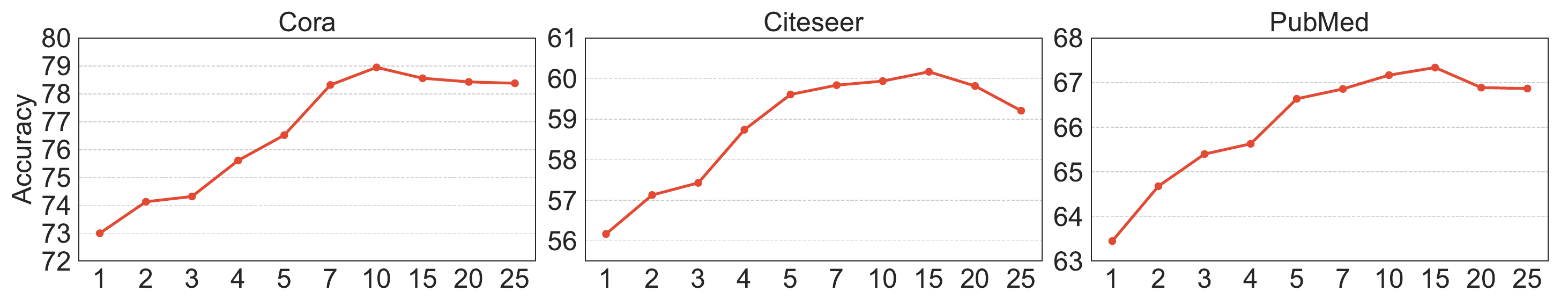} \label{subfig:budget}}\\
    \vspace{-0.4em}
    \caption{Hyper-parameters evaluation: (a) Alpha (b) Beta (c) Temperature (d) Budget Size}
    \vspace{-0.5cm}
    \label{fig:hyper}
\end{figure*}

\sloppy
We evaluate the impact of different hyper-parameters to answer \textbf{RQ2}. We evaluate our proposed model with different hyper-parameter settings: training size $\mathbf{N} \in \{C\times 1, C\times3, C\times5, C\times7\}$; the budget size $B \in [1,2,3,4,5,7,10,15,20,25]$ for graph-LLM based active learning with $N=C\times3$; the balance parameters $\alpha \in (0,0.5]$ with $\beta = 0.1$ and $\beta \in [0.01,0.03,0.05,0.1,0.2,0.3,0.4,0.5]$ with $\alpha = 0.3$, respectively; and the KD temperature $\tau \in [1,5]$. The evaluation results for training size are shown in Tables \ref{tab:compbaselinesCora}, \ref{tab:compbaselinesCiteseer}, and \ref{tab:compbaselinesPubMed}. The hyper-parameter evaluations are illustrated in Figure \ref{fig:hyper}.

\vspace{-0.2cm}
\begin{itemize}
    \item From Table \ref{tab:compbaselinesCora}, Table \ref{tab:compbaselinesCiteseer}, and Table \ref{tab:compbaselinesPubMed}, we can easily observe that when we increase the training size $\mathbf{N}$, the performance in terms of classification accuracy consistently improves. Especially when $\mathbf{N}$ is increased from $1$ to $3$, the performance significantly improves. 
    
    \item In Figure \ref{fig:hyper}(a), we observe that as the value of $\alpha$ increases, the performance initially improves, and then reaches a peak around $\alpha = 0.3$. However, the performance decreases drastically with further increases in $\alpha$ beyond this point.
    
    \item Figure \ref{fig:hyper}(b) illustrates that the performance increases when $\beta \in [0.01,0,1]$ and reaches the highest performance at $\beta = 0.1$. Then the performance keeps decreasing when we enlarge $\beta$, where the performance drops slightly when $\beta$ is less than 0.3, and it significantly drops after 0.3. The observed trend is quite understandable: as we increase the value of the balance parameters, the proportion of loss based on the ground truth gradually decreases. When the proportion falls below a certain threshold, the loss is primarily driven by the teacher loss and feature embedding loss. However, the quality of these pseudo-labels and feature embeddings generated from LLMs cannot be guaranteed. 
    
    \item Figure Figure \ref{fig:hyper}(c) shows the results of the evaluation on KD temperature $\tau$. It indicates that as we increase $\tau$, the performance initially experiences a significant improvement, stabilizes at a high level for $\tau \in [3, 5]$, and then drastically drops when $\tau$ changes from $5$ to $9$. It's not difficult to understand this trend: when $\tau$ is relatively small, the soft label probabilities distilled from the teacher model are informative and assist in optimizing the student model. However, when $\tau$ becomes large, the distilled knowledge becomes ambiguous, potentially leading to a smoothing effect on the student model's inference ability. 
    
    \item For the impact of budget size $B$, as shown in Figure \ref{fig:hyper}(d). As $B$ increases from 1 to 7, we observe a consistent improvement in performance. However, beyond this range, the performance remains stable or even drops, indicating an upper limit to the benefits gained from enlarging $B$. At this point, any further increase in the budget size results in higher costs, but only minimal gains in performance. This suggests that while increasing the budget can enhance performance up to a certain point, we need to make a trade-off between the performance and the cost.

\end{itemize}

\begin{table}[t]
\caption{Ablation study: Soft Labels (SL), Enhanced Rationales (ER), and Graph-LLM-based active learning (AL)}\label{tab:ablation}
\vspace{-0.3cm}
\centering
\footnotesize
    \begin{tabular}{|c|c|c||c|c|c|} 
        \hline 
         SL & ER & AL & Citeseer & Cora & PubMed   \\  
        \hline
        
         &  & &52.24±(1.04) &66.27±(5.04)  & 62.64±(0.85)\\
        \hline 
         &  &\checkmark &53.21±(2.25) &68.93±(3.88)  & 63.23±(1.15) \\
        \hline 
        \checkmark &  & &54.39±(3.84) &70.32±(3.81)  & 63.82±(0.99) \\
        \hline 
         & \checkmark & & 54.65±(2.58) & 68.87±(2.12) & 63.19±(2.41) \\
        \hline 
         &\checkmark  &\checkmark & 55.57±(2.18) & 71.12±(2.63)  & 64.33±(1.27) \\
        \hline 
         \checkmark & & \checkmark&55.41±(1.13) &72.91±(1.74)  & 64.59±(1.02) \\
        \hline 
         \checkmark & \checkmark &  &55.37±(1.83) &71.78±(2.80)  & 63.68±(0.82) \\
        \hline 
         \checkmark & \checkmark   &\checkmark & \textbf{57.43±(3.33)} & \textbf{74.32±(1.79)} & \textbf{65.40±(0.84)} \\
        \hline 
    \end{tabular}
    \vspace{-0.6cm}
\end{table}

\subsection{Ablation Study}
In this section, we design the ablation study to further investigate how different components contribute to the performance of our model and answer \textbf{RQ3}. Our model proceeds with LLMs and Graph-LLM-based active learning. For the LLMs, we further have two distinct perspectives: 1) soft labels and logits and 2) enhanced rationales. Thus, we investigate three components in our model design: soft labels and logits, enhanced rationales, and Graph-LLM-based active learning. (1) Soft labels and logits refer to the soft labels and logits that get from LLMs, which are used for knowledge distillation; (2) rationales refer to getting the enhanced explanation from LLMs, which will provide insights from the feature perspective at the embedding level. (3) Graph-LLM-based active learning refers to selecting valuable nodes for the model. The hyper-parameter is set to $N = k\times3$, $\alpha = 0.3$, $\beta = 0.1$, $T = 3$, $B = 3$, and backbone model is GCN. Note that when we solely apply active learning to the model, we select nodes with high confidence scores and prioritize the most important nodes. However, when we integrate the LLM into the model, we employ our Graph-LLM-based active learning strategy.

\begin{table}[t]
    \centering
    \footnotesize
    \caption{Ablation study: alignment and active learning}\label{tab:absMore}
    \vspace{-0.3cm}
    \tabcolsep=4pt
    \begin{tabular}{lccc}
    \toprule
        Strategy & Citeseer &  Cora & PubMed \\
        \midrule
        Alignment: max pooling & 52.28±(2.37) &70.88±(2.24) & 62.43±(1.21) \\
        AL: Random Selection & 54.97±(2.32) & 72.77±(1.86) & 63.58±(0.64)\\
        AL: w/o iteration & 55.84±(2.83) & 73.12±(1.97) & 64.43±(0.57) \\
        Our Model & \textbf{57.43±(3.33)} & \textbf{74.32±(1.79)} & \textbf{65.40±(0.84)} \\
    \bottomrule
    \end{tabular}
    \vspace{-0.6cm}
\end{table}

As illustrated in Table \ref{tab:ablation}, all these components contribute to the performance of our model. Among these components, the enhanced rationales have a relatively small impact on the performance. When we add soft-labels and active learning independently, the performance improves by a considerable margin. When we add these two components together, the performance has a significant improvement. The experimental results also showcase that LLMs can effectively enhance the performance of GNNs. Whether we incorporate the logits and enhanced rationales independently or combine them, there is a significant performance improvement. Furthermore, when all these components are integrated, our model achieves state-of-the-art performance.

For the rationale alignments, we further evaluate the two different alignment strategies: 1) max pooling and 2) our MLP-based alignment approach. For active learning, we evaluate different selection strategies: 1) randomly select nodes with pseudo-labels 2). select all valuable nodes at once with Graph-LLM-based AL; 2) select nodes in an iteration method with our Graph-LLM-based AL, which means we will select $b$ nodes per class until the total number of selected nodes reach $B\times C$. As shown in Table \ref{tab:absMore}, compared with the MLP-based alignment strategy, the improvement by using max pooling is limited, which is reasonable because max pooling will lose information during the pooling process. For active learning, the performance of the model with our Graph-LLM-based AL is better than random selection. Moreover, despite selecting all valuable nodes at once has shown significant performance improvement, the performance reaches a new high when we apply the iteration selection strategy. This enhancement is attributed to the iterative active learning process, where GNNs benefit from the LLM's zero-shot inference and reasoning ability to refine their predictions iteratively.


\section{Conclusion}
In this paper, we extend the task of node classification to a more challenging and realistic case where only few labeled data are available. To tackle this challenge, we propose a novel few-shot node classification model that leverages the zero-shot and reasoning ability of Large Language Models. We treat LLMs as a teacher to teach GNNs from two different perspectives including the logits from the distribution side and enhanced rationales from the feature side. Moreover, we proposed a Graph-LLM-based active learning method to further improve the generalizability of GNNs with few available data by actively selecting and distilling knowledge from LLMs. To assess the effectiveness of our model, extensive experiments have been conducted on three citation networks. The evaluation results demonstrate that our model achieves state-of-the-art performance and LLMs can effectively provide insights to GNNs from different perspectives, reaffirming its effectiveness in node classification, its superiority over baseline models, and its practical significance in addressing the challenges of few-shot node classification.

\bibliographystyle{IEEEtran}
\bibliography{ref}

\begin{thebibliography}{10}
\providecommand{\url}[1]{#1}
\csname url@samestyle\endcsname
\providecommand{\newblock}{\relax}
\providecommand{\bibinfo}[2]{#2}
\providecommand{\BIBentrySTDinterwordspacing}{\spaceskip=0pt\relax}
\providecommand{\BIBentryALTinterwordstretchfactor}{4}
\providecommand{\BIBentryALTinterwordspacing}{\spaceskip=\fontdimen2\font plus
\BIBentryALTinterwordstretchfactor\fontdimen3\font minus \fontdimen4\font\relax}
\providecommand{\BIBforeignlanguage}[2]{{%
\expandafter\ifx\csname l@#1\endcsname\relax
\typeout{** WARNING: IEEEtran.bst: No hyphenation pattern has been}%
\typeout{** loaded for the language `#1'. Using the pattern for}%
\typeout{** the default language instead.}%
\else
\language=\csname l@#1\endcsname
\fi
#2}}
\providecommand{\BIBdecl}{\relax}
\BIBdecl

\bibitem{chen2019semi}
W.~Chen, Y.~Gu, Z.~Ren, X.~He, H.~Xie, T.~Guo, D.~Yin, and Y.~Zhang, ``Semi-supervised user profiling with heterogeneous graph attention networks.'' in \emph{IJCAI}, vol.~19, 2019, pp. 2116--2122.

\bibitem{mohamed2020social}
A.~Mohamed, K.~Qian, M.~Elhoseiny, and C.~Claudel, ``Social-stgcnn: A social spatio-temporal graph convolutional neural network for human trajectory prediction,'' in \emph{CVPR}, 2020, pp. 14\,424--14\,432.

\bibitem{reau2023deeprank}
M.~R{\'e}au, N.~Renaud, L.~C. Xue, and A.~M. Bonvin, ``Deeprank-gnn: a graph neural network framework to learn patterns in protein--protein interfaces,'' \emph{Bioinformatics}, vol.~39, no.~1, p. btac759, 2023.

\bibitem{li2021dual}
R.~Li, H.~Chen, F.~Feng, Z.~Ma, X.~Wang, and E.~Hovy, ``Dual graph convolutional networks for aspect-based sentiment analysis,'' in \emph{ACL-IJCNLP}, 2021.

\bibitem{li2022distilling}
Q.~Li, X.~Li, L.~Chen, and D.~Wu, ``Distilling knowledge on text graph for social media attribute inference,'' in \emph{SIGIR}, 2022, pp. 2024--2028.

\bibitem{kipf2016semi}
T.~N. Kipf and M.~Welling, ``Semi-supervised classification with graph convolutional networks,'' \emph{arXiv preprint arXiv:1609.02907}, 2016.

\bibitem{abu2019mixhop}
S.~Abu-El-Haija, B.~Perozzi, A.~Kapoor, N.~Alipourfard, K.~Lerman, H.~Harutyunyan, G.~Ver~Steeg, and A.~Galstyan, ``Mixhop: Higher-order graph convolutional architectures via sparsified neighborhood mixing,'' in \emph{ICML}.\hskip 1em plus 0.5em minus 0.4em\relax PMLR, 2019, pp. 21--29.

\bibitem{hamilton2017inductive}
W.~Hamilton, Z.~Ying, and J.~Leskovec, ``Inductive representation learning on large graphs,'' \emph{NeurIPS}, vol.~30, 2017.

\bibitem{wu2020comprehensive}
Z.~Wu, S.~Pan, F.~Chen, G.~Long, C.~Zhang, and S.~Y. Philip, ``A comprehensive survey on graph neural networks,'' \emph{IEEE transactions on neural networks and learning systems}, vol.~32, no.~1, pp. 4--24, 2020.

\bibitem{wei2021few}
J.~Wei, C.~Huang, S.~Vosoughi, Y.~Cheng, and S.~Xu, ``Few-shot text classification with triplet networks, data augmentation, and curriculum learning,'' \emph{arXiv preprint arXiv:2103.07552}, 2021.

\bibitem{han2018fewrel}
X.~Han, H.~Zhu, P.~Yu, Z.~Wang, Y.~Yao, Z.~Liu, and M.~Sun, ``Fewrel: A large-scale supervised few-shot relation classification dataset with state-of-the-art evaluation,'' \emph{arXiv preprint arXiv:1810.10147}, 2018.

\bibitem{finn2017model}
C.~Finn, P.~Abbeel, and S.~Levine, ``Model-agnostic meta-learning for fast adaptation of deep networks,'' in \emph{ICML}, 2017, pp. 1126--1135.

\bibitem{ding2021few}
K.~Ding, Q.~Zhou, H.~Tong, and H.~Liu, ``Few-shot network anomaly detection via cross-network meta-learning,'' in \emph{WWW}, 2021.

\bibitem{ding2022meta}
K.~Ding, J.~Wang, J.~Caverlee, and H.~Liu, ``Meta propagation networks for graph few-shot semi-supervised learning,'' in \emph{AAAI}, 2022.

\bibitem{wan2021contrastive}
S.~Wan, Y.~Zhan, L.~Liu, B.~Yu, S.~Pan, and C.~Gong, ``Contrastive graph poisson networks: Semi-supervised learning with extremely limited labels,'' \emph{NeurIPS}, vol.~34, pp. 6316--6327, 2021.

\bibitem{zhu2021transfer}
Q.~Zhu, C.~Yang, Y.~Xu, H.~Wang, C.~Zhang, and J.~Han, ``Transfer learning of graph neural networks with ego-graph information maximization,'' \emph{NeurIPS}, vol.~34, pp. 1766--1779, 2021.

\bibitem{chen2022adversarially}
L.~Chen, X.~Li, and D.~Wu, ``Adversarially reprogramming pretrained neural networks for data-limited and cost-efficient malware detection,'' in \emph{SDM}.\hskip 1em plus 0.5em minus 0.4em\relax SIAM, 2022, pp. 693--701.

\bibitem{yao2021functionally}
H.~Yao, Y.~Wei, L.-K. Huang, D.~Xue, J.~Huang, and Z.~J. Li, ``Functionally regionalized knowledge transfer for low-resource drug discovery,'' \emph{NeurIPS}, vol.~34, 2021.

\bibitem{chen2023label}
Z.~Chen, H.~Mao, H.~Wen, H.~Han, W.~Jin, H.~Zhang, H.~Liu, and J.~Tang, ``Label-free node classification on graphs with large language models (llms),'' \emph{arXiv preprint arXiv:2310.04668}, 2023.

\bibitem{kojima2022large}
T.~Kojima, S.~S. Gu, M.~Reid, Y.~Matsuo, and Y.~Iwasawa, ``Large language models are zero-shot reasoners,'' \emph{NeurIPS}, 2022.

\bibitem{liu2023one}
H.~Liu, J.~Feng, L.~Kong, N.~Liang, D.~Tao, Y.~Chen, and M.~Zhang, ``One for all: Towards training one graph model for all classification tasks,'' \emph{arXiv preprint arXiv:2310.00149}, 2023.

\bibitem{he2023explanations}
X.~He, X.~Bresson, T.~Laurent, and B.~Hooi, ``Explanations as features: Llm-based features for text-attributed graphs,'' \emph{arXiv preprint arXiv:2305.19523}, 2023.

\bibitem{guo2024graphedit}
Z.~Guo, L.~Xia, Y.~Yu, Y.~Wang, Z.~Yang, W.~Wei, L.~Pang, T.-S. Chua, and C.~Huang, ``Graphedit: Large language models for graph structure learning,'' \emph{arXiv preprint arXiv:2402.15183}, 2024.

\bibitem{jiang2023structgpt}
J.~Jiang, K.~Zhou, Z.~Dong, K.~Ye, W.~X. Zhao, and J.-R. Wen, ``Structgpt: A general framework for large language model to reason over structured data,'' \emph{arXiv preprint arXiv:2305.09645}, 2023.

\bibitem{yu2023empower}
J.~Yu, Y.~Ren, C.~Gong, J.~Tan, X.~Li, and X.~Zhang, ``Empower text-attributed graphs learning with large language models (llms),'' \emph{arXiv preprint arXiv:2310.09872}, 2023.

\bibitem{velivckovic2017graph}
P.~Veli{\v{c}}kovi{\'c}, G.~Cucurull, A.~Casanova, A.~Romero, P.~Lio, and Y.~Bengio, ``Graph attention networks,'' \emph{arXiv preprint arXiv:1710.10903}, 2017.

\bibitem{li2018deeper}
Q.~Li, Z.~Han, and X.-M. Wu, ``Deeper insights into graph convolutional networks for semi-supervised learning,'' in \emph{AAAI}, vol.~32, no.~1, 2018.

\bibitem{hao2019collect}
F.~Hao, F.~He, J.~Cheng, L.~Wang, J.~Cao, and D.~Tao, ``Collect and select: Semantic alignment metric learning for few-shot learning,'' in \emph{ICCV}, 2019, pp. 8460--8469.

\bibitem{jiang2020multi}
W.~Jiang, K.~Huang, J.~Geng, and X.~Deng, ``Multi-scale metric learning for few-shot learning,'' \emph{IEEE Transactions on Circuits and Systems for Video Technology}, vol.~31, no.~3, pp. 1091--1102, 2020.

\bibitem{snell2017prototypical}
J.~Snell, K.~Swersky, and R.~Zemel, ``Prototypical networks for few-shot learning,'' \emph{NeurIPS}, vol.~30, 2017.

\bibitem{sung2018learning}
F.~Sung, Y.~Yang, L.~Zhang, T.~Xiang, P.~H. Torr, and T.~M. Hospedales, ``Learning to compare: Relation network for few-shot learning,'' in \emph{CVPR}, 2018, pp. 1199--1208.

\bibitem{zhou2019Meta}
F.~Zhou, C.~Cao, K.~Zhang, G.~Trajcevski, T.~Zhong, and J.~Geng, ``Meta-gnn: On few-shot node classification in graph meta-learning,'' in \emph{CIKM}, 2019, pp. 2357--2360.

\bibitem{g-meta}
K.~Huang and M.~Zitnik, ``Graph meta learning via local subgraphs,'' \emph{NeurIPS}, vol.~33, pp. 5862--5874, 2020.

\bibitem{sun2023large}
S.~Sun, Y.~Ren, C.~Ma, and X.~Zhang, ``Large language models as topological structure enhancers for text-attributed graphs,'' \emph{arXiv preprint arXiv:2311.14324}, 2023.

\bibitem{shen2017deep}
Y.~Shen, H.~Yun, Z.~C. Lipton, Y.~Kronrod, and A.~Anandkumar, ``Deep active learning for named entity recognition,'' \emph{arXiv preprint arXiv:1707.05928}, 2017.

\bibitem{bachman2017learning}
P.~Bachman, A.~Sordoni, and A.~Trischler, ``Learning algorithms for active learning,'' in \emph{ICML}.\hskip 1em plus 0.5em minus 0.4em\relax PMLR, 2017, pp. 301--310.

\bibitem{cai2017active}
H.~Cai, V.~W. Zheng, and K.~C.-C. Chang, ``Active learning for graph embedding,'' \emph{arXiv preprint arXiv:1705.05085}, 2017.

\bibitem{gao2018active}
L.~Gao, H.~Yang, C.~Zhou, J.~Wu, S.~Pan, and Y.~Hu, ``Active discriminative network representation learning,'' in \emph{IJCAI}, 2018.

\bibitem{wu2019active}
Y.~Wu, Y.~Xu, A.~Singh, Y.~Yang, and A.~Dubrawski, ``Active learning for graph neural networks via node feature propagation,'' \emph{arXiv preprint arXiv:1910.07567}, 2019.

\bibitem{hu2020graph}
S.~Hu, Z.~Xiong, M.~Qu, X.~Yuan, M.-A. C{\^o}t{\'e}, Z.~Liu, and J.~Tang, ``Graph policy network for transferable active learning on graphs,'' \emph{NeurIPS}, vol.~33, pp. 10\,174--10\,185, 2020.

\bibitem{feng2020graph}
W.~Feng, J.~Zhang, Y.~Dong, Y.~Han, H.~Luan, Q.~Xu, Q.~Yang, E.~Kharlamov, and J.~Tang, ``Graph random neural networks for semi-supervised learning on graphs,'' \emph{NeurIPS}, 2020.

\bibitem{reimers2019sentence}
N.~Reimers and I.~Gurevych, ``Sentence-bert: Sentence embeddings using siamese bert-networks,'' \emph{arXiv preprint arXiv:1908.10084}, 2019.

\bibitem{hinton2015distilling}
G.~Hinton, O.~Vinyals, J.~Dean \emph{et~al.}, ``Distilling the knowledge in a neural network,'' \emph{arXiv preprint arXiv:1503.02531}, vol.~2, no.~7, 2015.

\bibitem{wang2024distribution}
F.~Wang, T.~Zhao, and S.~Wang, ``Distribution consistency based self-training for graph neural networks with sparse labels,'' in \emph{WSDM}, 2024.

\bibitem{hassani2020contrastive}
K.~Hassani and A.~H. Khasahmadi, ``Contrastive multi-view representation learning on graphs,'' in \emph{ICML}.\hskip 1em plus 0.5em minus 0.4em\relax PMLR, 2020, pp. 4116--4126.

\end{thebibliography}


\end{document}